\documentclass[12pt]{article}
\usepackage[small,bf]{caption}
\usepackage{fullpage}
\usepackage{graphicx}
\usepackage{epstopdf}
\usepackage{amsmath}
\usepackage{csquotes}
\usepackage{url}
\usepackage{mathtools}
\usepackage{amssymb}
\usepackage{booktabs}
\usepackage{tikz}
\usepackage{xcolor}

\usepackage{pgfplots}
\pgfplotsset{compat=newest}

\usetikzlibrary{automata, trees, shapes, calc, patterns, angles, quotes, topaths}

\newcommand{\ones}{\mathbf 1}

\newcommand{\reals}{{\mbox{\bf R}}}

\newcommand{\Tr}{\mathop{\bf Tr}}
\newcommand{\diag}{\mathop{\bf diag}}

\newcommand{\prox}{\mathbf{prox}}

\newcommand{\argmin}{\mathop{\rm argmin}}

\newcommand{\eg}{{\it e.g.}}
\newcommand{\ie}{{\it i.e.}}

\newcommand{\BEAS}{\begin{eqnarray*}}
\newcommand{\EEAS}{\end{eqnarray*}}
\newcommand{\BEA}{\begin{eqnarray}}
\newcommand{\EEA}{\end{eqnarray}}
\newcommand{\BEQ}{\begin{equation}}
\newcommand{\EEQ}{\end{equation}}
\newcommand{\BIT}{\begin{itemize}}
\newcommand{\EIT}{\end{itemize}}

\newcounter{algorithmctr}[section]
\renewcommand{\thealgorithmctr}{\thesection.\arabic{algorithmctr}}
\newenvironment{algdesc}%
   {\refstepcounter{algorithmctr}\begin{list}{}{%
       \setlength{\rightmargin}{0\linewidth}%
       \setlength{\leftmargin}{.05\linewidth}}%
       \rmfamily\small
       \item[]{\setlength{\parskip}{0ex}\hrulefill\par%
        \nopagebreak{\bfseries\textsf{Algorithm \thealgorithmctr~}}}}%
   {{\setlength{\parskip}{-1ex}\nopagebreak\par\hrulefill} \end{list}}

\title{Eigen-Stratified Models}
\author{Jonathan Tuck \and Stephen Boyd}

\begin{document}
\maketitle

\begin{abstract}

Stratified models depend in an arbitrary way on a selected categorical 
feature that takes $K$ values, 
and depend linearly on the other $n$ features.
Laplacian regularization with respect to a graph on the feature values 
can greatly improve the performance of a stratified model, 
especially in the low-data regime.
A significant issue with Laplacian-regularized stratified models is that the model
is $K$ times the size of the base model, which can be quite large.

We address this issue by formulating \emph{eigen-stratifed models}, 
which are stratified models with an additional constraint that the model parameters 
are linear combinations of some modest number $m$ of bottom eigenvectors of the graph 
Laplacian, \ie, those associated with the $m$ smallest eigenvalues.
With eigen-stratified models,
we only need to store the $m$ bottom eigenvectors and
the corresponding coefficients as the stratified model parameters.
This leads to a reduction, sometimes large,
of model size when $m\leq n$ and $m \ll K$.
In some cases, the additional regularization implicit in
eigen-stratified models can improve out-of-sample performance over 
standard Laplacian regularized stratified models.
\end{abstract}

\newpage
\tableofcontents
\newpage

\section{Introduction}

\subsection{Eigen-stratified models}
Stratified models are models that depend in an arbitrary way on a selected
categorical feature (or set of features) that takes $K$ values,
and depend linearly on the other features.
For example in a date-stratified model 
we might have a different linear model for each day of the year, with $K=365$.
Laplacian regularization can be added to exploit some known relations 
among the categorical features, expressed as a graph.
In our date-stratified example, Laplacian regularization encourages the models
for adjacent dates to be close,
including the January 1 and December 31 models.
In this example, the underlying graph is a cycle with 365 vertices.

Laplacian regularization 
can greatly improve the performance of a stratified model, especially in the 
low-data regime.  In particular, it allows us to form a reasonable model even 
when we have no training data for some values of the categorical variable.
The number of parameters in a Laplacian-regularized stratified model is $K$,
the number of values of the categorical feature, times the size of the 
base model, which can be quite large when the catgorical values take
many values.  For example, a date-stratified model contains 365 times more
coefficients than the base model.
This is one of the challenges that we address in this paper.

Laplacian regularization encourages
the model parameters to vary smoothly across the graph that encodes our
prior information about the categorical values.
If the model parameters vary smoothly across the graph,
it is reasonable to assume they can be well approximated as linear combinations
of a modest number of the eigenvectors of the 
associated graph Laplacian associated with the smallest eigenvalues.
Our idea is simple:  We impose the constraint that the model parameters
are linear combinations of some number $m$ of the bottom eigenvectors of
the graph Laplacian.  
We refer to such a model as an \emph{eigen-stratified model}.
The resulting eigen-stratified model uses only
a factor $m$ parameters more than the
base model, compared to a factor $K$ for a general stratified model.
In addition to this savings in model size,
insisting that the model parameters are linear combinations of 
the bottom $m$ eigenvectors acts as an additional useful regularization,
that enforces smooth variation of the model
parameters across the graph.


In our date-stratified example, the bottom eigenvector is constant, and the next
ones occur in sine and cosine pairs, with periods one year, a half year,
one-third of year, and so on.
Using $m=7$, say, requires that the model parameters are Fourier 
series with $7$ terms (\ie, constant plus three harmonics).
So here the eigen-stratified model is very natural.

In more complex cases, the eigen-stratified model is far less obvious.
For example, the underlying graph can contain multiple edge weights,
which are hyper-parameters.  In any but the simplest cases,
we do not have analytical expressions for the eigenvalues and eigenvectors,
but they are readily computed, even for very large graphs.

\subsection{Related work}

\paragraph{Model approximations.}
It is quite common to approximate a larger model with a smaller, but only slightly less 
accurate model.
In signal processing, discrete signals are transformed into a basis where they may be
approximated by a linear combination of a small number of basis vectors, such as
complex exponentials or cosines, in order to achieve significant size compression at 
the cost of signal degradation, which in many cases is minimal 
\cite{ahmed1974dct,oppenheim2009dsp}.

\paragraph{Categorical embeddings.}
Learning low-dimensional vector representations of discrete variables is 
consistently used as a method to handle categorical features.
Embeddings are a popular tool in fields such as natural language processing,
to embed text as continuous vectors \cite{elman1990embeddings,mikolov2013word2vec,
guo2016entityembeddings}.
We can associate with each vertex the $m$ coefficients of the
bottom Laplacian eigenvectors.  This gives a Laplacian or spectral embedding 
of the features into $\reals^m$.

\paragraph{Spectral graph theory.}
The study of properties of graphs through their Laplacian eigen-decomposition is a 
long studied field in graph theory~\cite{chung1997spectral,cohensteiner2018spectral}.
Three example applications include spectral clustering \cite{ng2002spectral}, which is a form of
dimensionality reduction that uses the the eigen-decomposition of the graph Laplacian to cluster nodes 
in a graph;
finding the fastest Mixing Markov process on a graph, whose convergence guarantees rely on
the spectrum of the graph's Laplacian matrix (namely, the Fiedler eigenvalue of the Laplacian) 
\cite{boyd2004fmmp,sun2006fmmp,boyd2009fmmp};
and graph coloring \cite{brooks1941graphcoloring, brelaz1979graphcoloring}, where the goal is to assign 
one of a set of colors to a graph node such that no two adjacent nodes share a color.
Graph coloring is an NP-hard task in general, but ideas from spectral graph theory 
are naturally used as heuristics to sub-optimally solve this problem \cite{aspvall1984graphcoloring}.

\paragraph{Laplacian regularization in large-scale optimization.}
There are many general methods to solve convex optimization problems
with Laplacian regularization. 
Examples include the alternating direction method of multipliers (ADMM) 
\cite{tuck2019stratmodels},
majorization-minimization (MM) \cite{tuck2018distributed},
and Anderson accelerated Douglas-Rachford splitting \cite{fu2019a2dr}.
In addition, the idea of applying Laplacian approximations to large-scale optimization 
problems has been studied in the past, where one approximates the graph Laplacian 
by a linear combination of the eigenvectors to solve extremely large 
semidefinite programs 
in, \eg, maximum variance unfolding~\cite{weinberger2007laplacian}.

\subsection{Outline}
In~\S\ref{ss:strat_models} we review stratified models, fixing our notation;
in~\S\ref{ss:eigen_strat_models} we formally describe the eigen-stratified model fitting 
problem, and in~\S\ref{s-distr-method}, we give a distributed solution method.
In~\S\ref{s-examples} we give some simple numerical examples, 
carried out using an accompanying open-source implementation of our method.

\section{Eigen-stratified models}\label{s:eigen_strat}
In this section, we give a brief overview of stratified models;
see \cite{tuck2019stratmodels} for much more detail.

\subsection{Stratified models}\label{ss:strat_models}
We fit a model to data records of the form 
$(z,x,y) \in \mathcal Z \times \mathcal X \times \mathcal Y$.
Here $z\in \mathcal Z$ is the feature over which we stratify, $x\in \mathcal X$ is the 
other features, and $y\in \mathcal Y$ is the outcome, label, or dependent variable.
The feature and label spaces $\mathcal X$ and $\mathcal Y$ 
are arbitrary data types; the stratified feature values 
$\mathcal Z$, however, must consist of only $K$ possible values, which we denote as 
$\mathcal Z =\{1, \ldots, K\}$.

A stratified model is built on top of a base model, which models pairs $(x,y)$
(or, when $x$ is absent, just $y$).
The base model is parametrized by a parameter vector $\theta \in \Theta \subseteq \reals^n$.
In a stratified model, we use a different value of the parameter $\theta$ for each value of 
$z$.
We denote these parameters as $\theta_1, \ldots, \theta_K$, where $\theta_k$ is the parameter
value used when $z=k$.
We let $\theta \in \reals^{n \times K}$ denote the parameter values for the stratified model,
where
\[
  \theta = [\theta_1 \cdots \theta_K] \in \reals^{n \times K}.
\]
(In \cite{tuck2019stratmodels}, the individual parameter vectors 
$\theta_k$ were stacked into one vector of dimension $nK$; 
here it will be more convenient to assemble them into a matrix.)

To choose the parameters $\theta_1, \ldots, \theta_K$, we minimize
\BEQ
\sum_{k=1}^K \left( \ell_k(\theta_k) +
r (\theta_k) \right) +
\mathcal L(\theta).
\label{eq:strat-obj}
\EEQ
The first term is the sum of $K$ local objective functions, with the 
$k$th local objective function consisting of a local loss of the form
\BEQ\label{eq:loc-obj}
\ell_k(\theta) = \sum_{i:z_i=k} l(\theta,x_i,y_i),
\EEQ
with loss function $l: \Theta \times \mathcal X \times \mathcal Y \to \reals$,
and local regularizer $r: \Theta \to \reals \cup \{\infty\}$.
(Infinite values of the regularizer encode constraints on allowable into a matrix.)
Choosing $\theta_k$ to minimize $\ell_k(\theta_k)+r(\theta_k)$ gives the
regularized empirical risk minimization model parameters, based only on the
data records that take the particular value of the stratification feature $z=k$.

The second term $\mathcal L(\theta)$ in~(\ref{eq:strat-obj}) 
measures the non-smoothness of the model parameters over $z \in \mathcal Z$.
Let $W \in \reals^{K\times K}$ be a symmetric matrix with nonnegative entries.
The associated \emph{Laplacian regularization} or \emph{Dirichlet energy} is the function
$\mathcal L: \reals^{n \times K} \to \reals$ given by
\BEQ\label{eq:lapl}
\mathcal L(\theta)
= \frac{1}{2}\sum_{i,j=1}^K W_{ij} \|\theta_i - \theta_j\|_2^2.
\EEQ
We can associate the Laplacian regularization with a graph with
$K$ vertices, with an edge $(i,j)$ for each positive $W_{ij}$, with weight $W_{ij}$.
We can express the Laplacian regularization as the positive semidefinite quadratic form
\[
\mathcal L(\theta) = (1/2) \Tr(\theta L \theta^T),
\]
where $L \in \reals^{K \times K}$ is the (weighted) Laplacian matrix associated 
with the weighted graph, given by
\[
L_{ij}  = \left\{ \begin{array}{ll}
-W_{ij} & i \neq j \\
\sum_{k=1}^K W_{ik} & i=j
\end{array} \right.
,\qquad i, j = 1, \ldots, K.
\]
We note that the Laplacian regularization $\mathcal L(\theta)$ is separable 
in the rows of $\theta$.

We refer to the model obtained by 
solving \eqref{eq:strat-obj} as a \emph{standard stratified model}.
When the loss function $\ell$ and local regularization function $r$ 
are convex, the objective in \eqref{eq:strat-obj} is convex,
which implies that a global solution can be found efficiently
\cite{boyd2004convex}.
When this assumption does not hold, heuristic methods can be 
used to approximately solve \eqref{eq:strat-obj}.

\subsection{Eigen-stratified models}\label{ss:eigen_strat_models}
The eigen-decomposition of the Laplacian matrix $L$ is
\[
L = Q \Lambda Q^T,
\]
where $\Lambda \in \reals^{K \times K}$, a diagonal matrix consisting of the eigenvalues of $L$, 
is of the form $\Lambda = \diag(\lambda_1, \ldots, \lambda_K)$ with 
$0 = \lambda_1 \leq \cdots \leq \lambda_K$,
and $Q = (q_1, \ldots, q_{K}) \in \reals^{K \times K}$ is a matrix of orthonormal 
eigenvectors of $L$.
Since $L\ones =0$, where $\ones$ is the vector with all entries one,
we have $\lambda_1 = 0$, and $q_1 = \ones/\sqrt K$ \cite{spielman2010algorithms}.
(When the graph is connected, $q_1$ is unique, and $\lambda_2>0$.)
In many cases, the eigenvectors and eigenvalues of a graph Laplacian matrix can be computed
analytically; in Appendix~\ref{app:graphs}, we mention a few of these common graphs and give
their eigenvectors and eigenvalues.

For $m \leq K$, we refer to $\lambda_1, \ldots, \lambda_m$ as the bottom $m$ eigenvalues,
and $q_1, \ldots, q_m$ as the bottom $m$ eigenvectors.
They are an orthonormal basis of the subspace of $\reals^K$ that is smoothest, \ie,
minimizes $\Tr \tilde Q^T L \tilde Q$,
where $\tilde Q =[q_1~\cdots~q_{m}] \in \reals^{K \times m}$,
subject to $\tilde Q^T\tilde Q = I_m$.
Roughly speaking, functions on $\mathcal Z$ that are smooth should be well 
approximated by a linear combination of the bottom $m$ eigenvectors (for suitable $m$).

Assuming that $\theta$ has low Dirichlet energy, \ie, 
a small Laplacian regularization term, we conclude that its
rows are well approximated by a linear combination of the bottom $m$ eigenvectors.
This motivates us to impose a further constraint on the rows of $\theta$: They
must be linear combinations of the bottom $m$ eigenvectors of $L$.
This can be expressed as
\BEQ
\label{eq:approximation}
\theta = Z {\tilde Q}^T,
\EEQ
where $Z \in \reals^{n \times m}$ are the (factorized) model parameters
and $\tilde Q \in \reals^{K \times m}$ are the bottom $m$ eigenvectors of $L$.

Adding the constraint \eqref{eq:approximation} to the Laplacian regularized 
stratified model fitting problem \eqref{eq:strat-obj}, we obtain the problem
\BEQ
\label{eq:strat_cons}
\begin{array}{ll}
\mbox{minimize} & \sum_{k=1}^K (\ell_k(\theta_k) + r(\theta_k)) + \mathcal L(\theta)\\
\mbox{subject to} & \theta = Z {\tilde Q}^T ,
\end{array}
\EEQ
where now both $\theta$ and $Z$ are optimization variables, coupled
by the equality constraint.
We can express the Laplacian regularization term in \eqref{eq:strat_cons}
directly in terms of $Z$ as
\[
\mathcal L(\theta) =
(1/2) \Tr(\theta L \theta^T) 
= (1/2) \Tr(Z {\tilde Q}^T L {\tilde Q} Z^T) 
= (1/2) \|Z \Lambda_m^{1/2}\|^2,
\]
where $\Lambda_m = \diag(\lambda_1, \ldots, \lambda_m)$ is the diagonal matrix
of the eigenvalues corresponding to the bottom $m$ eigenvectors of $L$.
We refer to the model obtained by solving \eqref{eq:strat_cons} as 
an \emph{eigen-stratified model}.

We note that the sum of empirical losses and local regularization are clearly
separable in the columns of $\theta$, 
and the Laplacian regularization is a separable function in the rows of $Z$.

\paragraph{Comparison to standard stratified models.}
With standard stratified models, we allow arbitrary variations of the model parameter $\theta$
across the graph.
With eigen-stratified models, we sharply limit how $\theta$ varies across the graph by constraining
$\theta$ to be a linear combination of the $m$ bottom eigenvectors of
the graph.

\paragraph{Storage.}
The standard stratified model requires us to store $Kn$ model parameters.
An eigen-stratified model, on the other hand, stores $m(K+n)$ variables in the 
eigenvectors $\tilde Q$ and the factorized model parameters $Z$. 
This implies that when $m \leq n$ and $m \ll K$, the storage savings is significant.

\paragraph{Convexity.}
If the $\ell_k$ and $r$ are convex, then \eqref{eq:strat_cons} is a convex problem,
which is readily solved globally in an efficient manner.
It is easily formulated using domain specific languages for convex
optimization \cite{boyd2004convex,grant2006dcp,grant2014cvx,diamond2016cvxpy,fu2019cvxr}.
If any of the $\ell_k$ or $r$ are nonconvex, it is a hard problem to solve 
\eqref{eq:strat_cons} globally.
In this case, our method (described in \S\ref{s-distr-method})
will provide a good heuristic approximate solution.

\paragraph{The two extremes.}
For a given set of edge weights, we analyze the behavior of the eigen-stratified model as we vary $m$.
When we take $m = 1$ and the graph is connected, 
we recover the common model (\ie, a stratified model with all $\theta_i$ equal).
We can see this by noting that when $m=1$ and the graph is connected,
the constraint in~\eqref{eq:strat_cons} becomes a 
consensus constraint (recall that the bottom eigenvector of a Laplacian matrix is a scalar multiple of $\ones$).
If we take $m =K$, the eigen-stratified model is the same as the 
standard stratified model.

\section{Distributed solution method}
\label{s-distr-method}
In this section we describe a distributed algorithm
for solving the fitting problem \eqref{eq:strat_cons}.
To derive the algorithm, we first express \eqref{eq:strat_cons} in the equivalent form 
\BEQ
\label{eq:variant}
\begin{array}{ll}
\mbox{minimize} & \sum_{k=1}^K (\ell_k(\theta_k) + r({\tilde\theta}_k)) + (1/2) \|Z \Lambda_m^{1/2}\|^2\\
\mbox{subject to} & \theta = Z {\tilde Q}^T \quad \theta = {\tilde\theta},
\end{array}
\EEQ
where we have introduced an additional
optimization variable $\tilde\theta\in \reals^{n \times K}$.

The augmented Lagrangian $L_{\rho}$ of \eqref{eq:variant} has the form
\BEAS
L_{\rho}(\theta, \tilde{\theta}, Z, u, \tilde u) 
&=& \sum_{k=1}^K (\ell_k(\theta_k) + r({\tilde\theta}_k)) 
+ (1/2) \|Z\Lambda_m^{1/2}\|^2\\
&& \mbox{}  + (1/2\rho)\|\theta - \tilde{\theta} + u\|_2^2
+ (1/2\rho)\|\tilde{\theta} - Z{\tilde Q}^T + \tilde u\|_2^2,
\EEAS
where $u\in\reals^{n \times K}$ and $\tilde u\in\reals^{n \times K}$
are the (scaled) dual variables associated 
with the two constraints in \eqref{eq:variant},
respectively, and $\rho>0$ is the penalty parameter.
The ADMM algorithm (in scaled dual form) for
the splitting $(\theta,Z)$ and
${\tilde\theta}$ consists of the iterations
\BEAS
\theta^{i+1}, Z^{i+1}
&\coloneqq&
\argmin_{\theta,Z}
L_{\rho}(\theta, {\tilde\theta}^{i+1}, Z, u^i, \tilde u^i) \\
{\tilde\theta}^{i+1} &\coloneqq&
\argmin_{\tilde\theta} L_{\rho}(\theta^{i+1}, {\tilde\theta}, Z^{i+1}, u^i, \tilde u^i)\\
u^{i+1} &\coloneqq& u^i + \theta^{i+1} - {\tilde\theta}^{i+1} \\
\tilde u^{i+1} &\coloneqq& {\tilde{u}}^i + {\tilde\theta}^{i+1} - Z^{i+1}{\tilde Q}^T.
\EEAS
If the $\ell_k$ and $r$ are convex, the iterates $\theta^i$, $\tilde\theta^i$ are 
guaranteed to converge to each other and 
$\theta^i$, $\tilde\theta^i$, and $Z^i$ are guaranteed to converge to a primal optimal 
point of \eqref{eq:variant} \cite{boyd2011distributed}.

This algorithm can be greatly simplified (and parallelized) by making use of a few observations.
Our first observation is that the first step in ADMM can be expressed as
\[
\theta^{i+1}_k = \prox_{\rho l_k}(\tilde\theta_k^i-u_k^i), \quad
k=1,\ldots,K,
\]
where $\prox_{g}: \reals^n \to \reals^n$ is the
proximal operator of the function $g$ \cite{PB:14},
and
\[
Z^{i+1} = (1/\rho) ({\tilde u}^i + {\tilde \theta}^i) {\tilde Q} (\Lambda_m + (1/\rho)I)^{-1}.
\]
This means that we can compute $\theta^{i+1}$
and $Z^{i+1}$ at the same time, since they do not depend on each other.
Also, we can compute $\theta_1^{i+1},\ldots,\theta_K^{i+1}$ in parallel.

Our second observation is that the second step in ADMM can be expressed as 
\[
\tilde\theta^{i+1}_k = \prox_{\rho r}(Z^{i+1} {\tilde Q}^T - {\tilde u^i}), \quad
k=1,\ldots,K,
\]
Similarly, we can compute ${\tilde\theta}_1^{i+1},\ldots,{\tilde\theta}_K^{i+1}$ 
in parallel.

Combining these observations leads to Algorithm~\ref{alg:eigen_strat}.

\begin{algdesc}
\label{alg:eigen_strat}
\emph{Distributed method for fitting eigen-stratified models.}
\begin{tabbing}
    {\bf given} Loss functions $\ell_1,\ldots,\ell_K$, 
    local regularization function $r$, penalty parameter $\rho>0$,\\
    $m$ bottom eigenvectors of the graph Laplacian matrix $\tilde Q \in \reals^{K \times m}$,\\
    and diagonal matrix with corresponding $m$ bottom eigenvalues $\Lambda_m \in \reals^{K \times K}$.\\
    \emph{Initialize}.
    ${\tilde\theta}^0=u^0={\tilde u}^0=0$.\\
    {\bf repeat} \\
    \qquad \=\ \bf{in parallel}\\
    \qquad \qquad \=\
    \emph{Evaluate proximal operator of $\ell_k$.}
    $\theta^{i+1}_k = \prox_{\rho \ell_k}(\tilde\theta_k^i-u_k^i), 
    \quad k = 1,\ldots,K$\\
    \qquad \=\
    \emph{Update $Z$.}
    $Z^{i+1} = (1/\rho) ({\tilde u}^i + {\tilde \theta}^i) {\tilde Q} (\Lambda_m + (1/\rho)I)^{-1}$\\

    \qquad \=\ \bf{in parallel}\\
    \qquad \qquad \=\
    \emph{Evaluate proximal operator of $r$.}
    $\tilde\theta^{i+1}_k = \prox_{\rho r}(Z^{i+1} {\tilde Q}^T - {\tilde u^i}), 
    \quad k = 1,\ldots,K$\\
    \qquad \=\
    \emph{Update the dual variables.}
    $u^{i+1} \coloneqq u^i + \theta^{i+1} - {\tilde\theta}^{i+1};
    \quad \tilde u^{i+1} \coloneqq {\tilde{u}}^i + {\tilde\theta}^{i+1} - Z^{i+1}{\tilde Q}^T$\\
    {\bf until convergence}
\end{tabbing}
\end{algdesc}

\paragraph{Complexity.}
Generally, the dominant cost of this algorithm depends on the
complexity of computing a single proximal operator of $l_k$ or $r$.
Otherwise, the dominant costs are in multiplying a dense $n \times K$ matrix,
a dense $K \times m$ matrix, and a diagonal $K \times K$ matrix together.

\section{Examples}\label{s-examples}

In this section, we illustrate the efficacy of the proposed method
on two simple and relatively small examples.

\paragraph{Software implementation.}
An implementation of our method for fitting an eigen-stratified model is given
as an extension of the stratified model fitting implementation 
in \cite{tuck2019stratmodels},
available at \url{www.github.com/cvxgrp/strat_models} 
(along with the accompanying examples).
To fit an eigen-stratified model, one may invoke
\[
  \verb|model.fit(data, num_eigen=None, ...)|.
\]
Here, $\verb|data|$ are the problem data (\ie, $(z, x, y)$ or $(z, y)$) and 
$\verb|num_eigen|$ is the number of bottom eigenvectors to use in the eigen-stratified model
(\ie, $m$); 
if $\verb|num_eigen|$ is $\verb|None|$, a standard Laplacian-regularized stratified 
model is fit.

\subsection{Cardiovascular disease prediction}\label{ss:cardio}
We consider the problem of predicting whether a patient has cardiovascular disease,
given their sex, age, and other medical features.

\paragraph{Dataset.}
We use data describing approximately 70000 patients across the world 
\cite{kaggle2019cardiodata}.
The dataset is comprised of males and females between the ages of 39 and 65 (inclusive),
with approximately 50\% of the patients diagnosed with cardiovascular disease.

There are 9 raw medical features in this dataset, which include:
height, weight, systolic blood pressure (a categorical feature with values 
``below average", ``average", and ``above average"), 
diastolic blood pressure (a categorical feature with values 
``below average", ``average", and ``above average"), cholesterol,
glucose levels, whether or not the patient smokes, whether or not the patient drinks alcohol,
and whether or not the patient undergoes regular physical activity.
We randomly partition the data into a training set consisting of 5\% of the records, 
a validation set containing 5\% of the records,
and a test set containing the remaining 90\% of the records.
We choose extremely small training and validation sets
to illustrate the efficacy of stratified models in low-data regimes.

\paragraph{Data records.}
We performed basic feature engineering on the raw medical features to 
derive a feature vector $x \in \reals^{14}$ (\ie, $n=14$), 
namely scalarization and converting the systolic blood pressure
and diastolic blood pressure basic categorical features into multiple features via one-hot encoding,
and adding a constant feature.
The outcomes $y \in \{0,1\}$ denote whether or not the patient has contracted cardiovascular
disease, with $y=1$ meaning the patient has cardiovascular disease.
The stratification feature $z$ is a tuple of the patient's sex and age; \eg,
$z=(\mathrm{Male}, 47)$ corresponds to a 47 year old male.
The number of stratification feature values is thus $K = 2 \cdot 27 = 54$.

\paragraph{Data model.}
We model the conditional probability of contracting cardiovascular disease given
the features using a logistic regression base model (with intercept).
We use logistic loss and sum of squares regularization,
\ie, $r = (1/2)\|\cdot\|_2^2$, with associated 
hyper-parameter $\gamma_{\mathrm{local}}$.

\begin{figure}
  \centering
  \begin{tikzpicture}
  \def \mh {4}
  \def \fh {2}
  \def \minrad {4em}
      \node[shape=circle,draw=black, minimum size=\minrad] (M1) at (0,\mh) {M,39};
      \node[shape=circle,draw=black, minimum size=\minrad] (F1) at (0,\fh) {F,39};
      \node[shape=circle,draw=black, minimum size=\minrad] (M2) at (3,\mh) {M,40};
      \node[shape=circle,draw=black, minimum size=\minrad] (F2) at (3,\fh) {F,40};
      \draw[-] (M1) -- (F1);
      \draw[-] (M2) -- (F2);    
      \draw[-] (M1) -- (M2);
      \draw[-] (F1) -- (F2);
  
      \node[] (dotsM) at (5,\mh) {$\cdots$};
      \node[] (dotsF) at (5,\fh) {$\cdots$};
  
      \node[shape=circle,draw=black, minimum size=\minrad] (M99) at (7,\mh) {M,64};
      \node[shape=circle,draw=black, minimum size=\minrad] (F99) at (7,\fh) {F,64};
      \node[shape=circle,draw=black, minimum size=\minrad] (M100) at (10,\mh) {M,65};
      \node[shape=circle,draw=black, minimum size=\minrad] (F100) at (10,\fh) {F,65};
      \draw[-] (M99) -- (F99);
      \draw[-] (M100) -- (F100);    
      \draw[-] (M99) -- (M100);
      \draw[-] (F99) -- (F100);
  
      \draw[-] (dotsM) -- (M99);
      \draw[-] (dotsM) -- (M2);
      \draw[-] (dotsF) -- (F99);
      \draw[-] (dotsF) -- (F2);
  \end{tikzpicture}
  \caption{Regularization graph for \S\ref{ss:cardio}.}
  \label{fig:cardio_reg_graph}
  \end{figure}
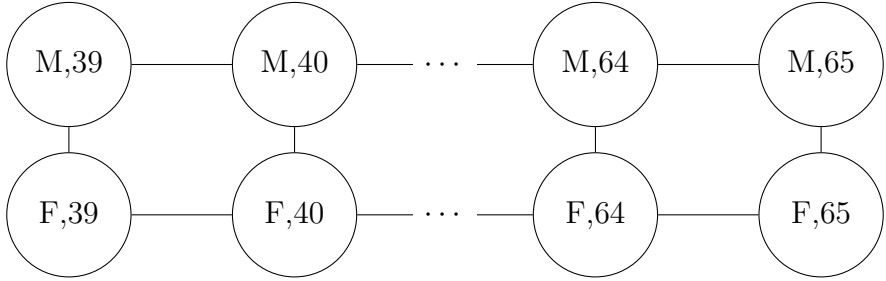

\paragraph{Regularization graph.}
We take the Cartesian product of two regularization graphs:
\begin{itemize}
\item \emph{Sex.} The regularization graph is a path graph that has one edge between 
male and female, with edge weight $\gamma_{\mathrm{sex}}$.
\item \emph{Age.} The regularization graph is a path graph between ages, with edge
weight $\gamma_{\mathrm{age}}$.
\end{itemize}
Figure~\ref{fig:cardio_reg_graph} illustrates the structure of this 
regularization graph.
  
From Appendix~\ref{app:graphs}, the eigenvectors of the regularization graph's Laplacian,
$q_{i,j}$ for $i = 1, 2$ and $j = 1, \ldots, 27$, are given in closed-form as
\BEAS
&q_{i,j} = {\tilde q}_{i,j} / \|{\tilde q}_{i,j}\|_2,\\
&{\tilde q}_{i,j} =  \cos((\pi/2)(i-1)(v-1/2)) \otimes \cos((\pi/27) (j-1) (v - 1 / 2))
  \quad i = 0, 1, \quad j = 0, \ldots, 26,
\EEAS
where $\otimes$ denotes a Kronecker product, $v = (0, \ldots, K-1)$, and $\cos(\cdot)$ is applied 
elementwise. (It is convenient for the eigenvectors to be indexed by two numbers, 
corresponding to the sex and age subgraphs that make up the regularization graph.)

\begin{figure}
  \centering
    \includegraphics[width=\textwidth]{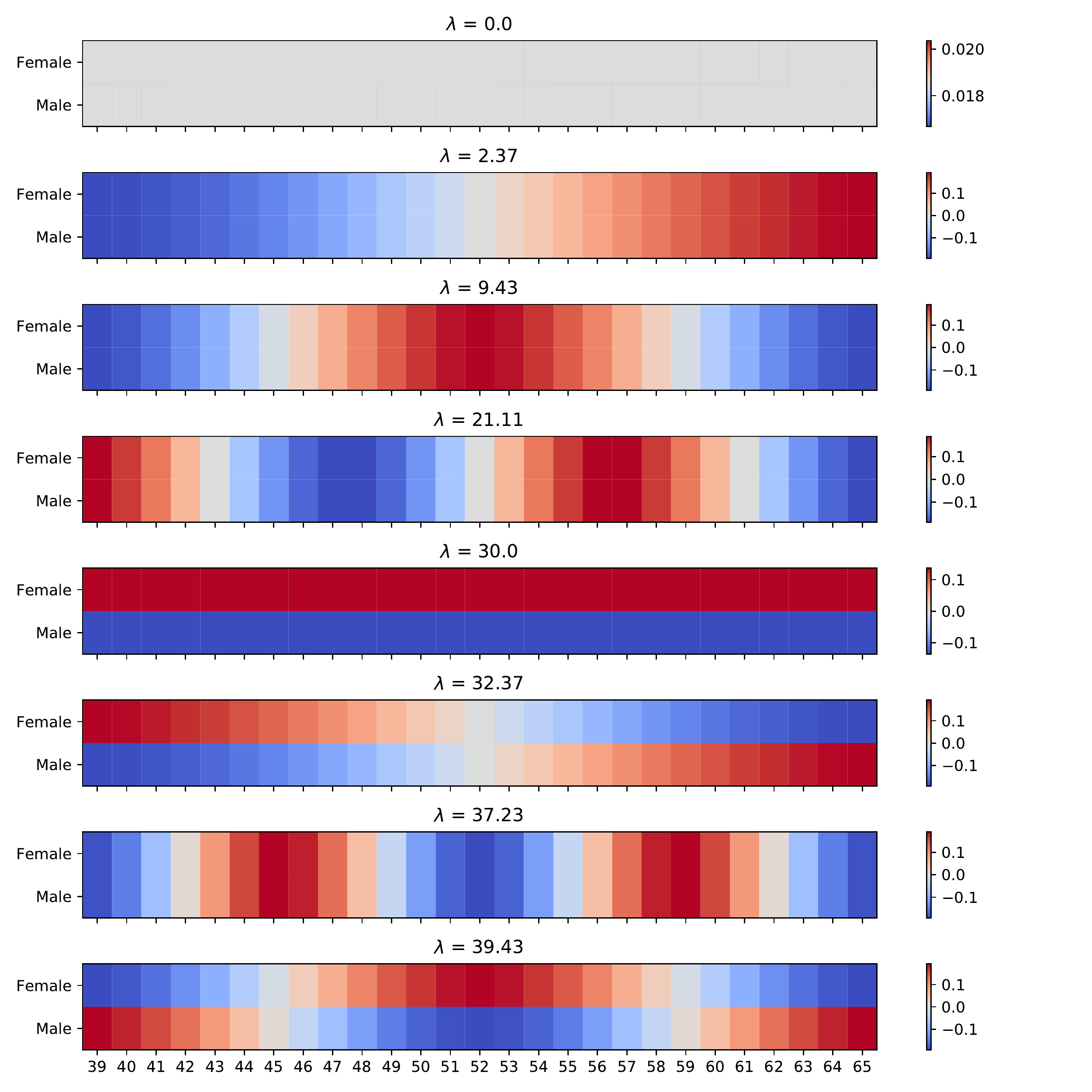}
      \caption{Heatmaps of the eigenvectors of the sex/age regularization 
      graph Laplacian corresponding to the bottom 8 eigenvalues of the Laplacian.}
  \label{fig:cardio_eigen_heatmap}
\end{figure}

Figure \ref{fig:cardio_eigen_heatmap} plots 8 of the 54 eigenvectors of the sex/age regularization 
graph Laplacian, with the particular sex/age edge weights 
$(\gamma_{\mathrm{sex}},\gamma_{\mathrm{age}}) = (15,175)$, 
sorted in increasing order corresponding to the bottom 8 eigenvalues of the Laplacian.

\paragraph{Results.}
For each of the fitting methods, we ran a crude hyper-parameter search over their hyper-parameters
and selected hyper-parameters that performed well over the validation set.
For the separate model, we used $\gamma_{\mathrm{local}}=35$, and
for the common model, we used $\gamma_{\mathrm{local}}=5$.
(Recall that the separate model is a stratified model with all edge weights zero, 
and a common model is a stratified model with all edge weights $+\infty$ \cite{tuck2019stratmodels}.)
For the standard stratified model, we used $\gamma_{\mathrm{local}}=0.01$,
$\gamma_{\mathrm{sex}} = 125$ and $\gamma_{\mathrm{age}} = 150$.
For the eigen-stratified model, we used $\gamma_{\mathrm{local}}=2.5$,
$\gamma_{\mathrm{sex}} = 15$ and $\gamma_{\mathrm{age}} = 175$,
and $m=5$.
Table~\ref{tab:cardio} shows the average negative log likelihood (ANLL)
over the training, validation, and test datasets 
for the separate, common, standard stratified and eigen-stratified models.
\begin{table}
  \caption{Results for \S\ref{ss:cardio}.}
    \vspace{.4em}
    \centering
    \begin{tabular}{llll}
      \toprule
      Model & Train ANLL & Validation ANLL & Test ANLL \\
      \midrule
      Separate & 0.607 & 0.656 & 0.658\\
      Common & 0.610 & 0.597 & 0.615\\
      Standard stratified & 0.572 & 0.567 & 0.597\\
      Eigen-stratified & 0.581 & 0.563 & 0.596\\
      \bottomrule
    \end{tabular}
    \label{tab:cardio}
  \end{table}
We see that this test ANLL attains a minimum when only $5$ bottom eigenvectors are used 
for the eigen-stratified model. 
This minimum test ANLL of the eigen-stratified model is competitive with 
(in fact, slightly smaller than) the test ANLL of the standard stratified model.

In the eigen-stratified model with $m=5$, the model parameters are linear combinations of 
$5$ bottom eigenvectors.
There are $n K = 14\cdot54=756$ parameters in the standard stratified 
model to store, whereas the eigen-stratified model with minimum test ANLL stores 
$m(n+K) = 5\cdot(14+54)=340$ values, or approximately 45\% as many parameters.
So there is some storage efficiency gain even in this very simple example.

\subsection{Weather distribution modeling}\label{ss:weather}
We consider the problem of modeling the distribution of weather temperature as a function of
week of year and hour of day.

\paragraph{Data records and dataset.}
We use temperature measurements from the city of Atlanta, Georgia for all of 
2013 and 2014, sampled every hour (for a total of approximately 17500 measurements).
The temperature is in Celsius; we round the temperatures to the nearest integer.
There are $n=43$ unique temperatures, ranging from -9 to 33 Celsius.
Each data record includes the temperature, as well as the week of the year and the hour 
of the day (which will be the stratification features).
The number of stratification features is $K=52 \cdot 24 = 1248$.

We partition the dataset into three separate sets; a training set consisting of 30\% of the data, 
a validation set consisting of 35\% of the data, 
and a held-out test set consisting of the remaining 35\% of the data.
The model is trained on approximately 4.2 samples per stratification feature.

\paragraph{Data model.}
We model the distribution of temperature in Atlanta at each week and hour using a 
non-parametric discrete distribution~\cite{tuck2019stratmodels}.
Our local regularizer is a sum of two regularizers: a sum of squares regularizer and a 
scaled sum of squares regularizer on the difference between adjacent parameters, \ie, 
$r(\theta) = \gamma_1 r_1(\theta) + \gamma_2 r_2(\theta)$, with $r_1(\theta) = (1/2)\|\theta\|_2^2$ 
and $r_2(\theta) = (1/2) \sum_{i=1}^{n} (\theta_{i+1}-\theta_i)^2$;
the associated hyper-parameters with each are $\gamma_1$ and $\gamma_2$.
The distribution $p_z$ at each node $z$ is calculated as
\[
  p_z = \frac{\exp(\theta_z)}{\sum_{i=1}^n \exp(\theta_z)_i},
\]
where $\exp(\cdot)$ is evaluated elementwise.

\begin{figure}
  \centering
\begin{tikzpicture}
  \begin{axis}[
  axis equal image,
  hide axis,
  z buffer = sort,
  view = {122}{30},
  scale = 2.5
  ]
  \addplot3[
  surf,
  line width=0.3pt,
  faceted color=black,
  shader = faceted interp,
  colormap/blackwhite,
  samples = 24,
  samples y = 52,
  domain = 0:2*pi,
  domain y = 0:2*pi,
  point meta=x+3*z*z-0.25*y
  ](
  {(3+sin(deg(\x)))*cos(deg(\y))},
  {(3+sin(deg(\x)))*sin(deg(\y))},
  {cos(deg(\x))}
  );
  \end{axis}
  \end{tikzpicture}
  \caption{Regularization graph for \S\ref{ss:weather}. 
  Each node corresponds to a week of year and hour of day, where
  the toroidal direction corresponds to increasing week of year, 
  and the poloidal direction corresponds to increasing hour of day.}
  \label{fig:weather_reg_graph}
\end{figure}
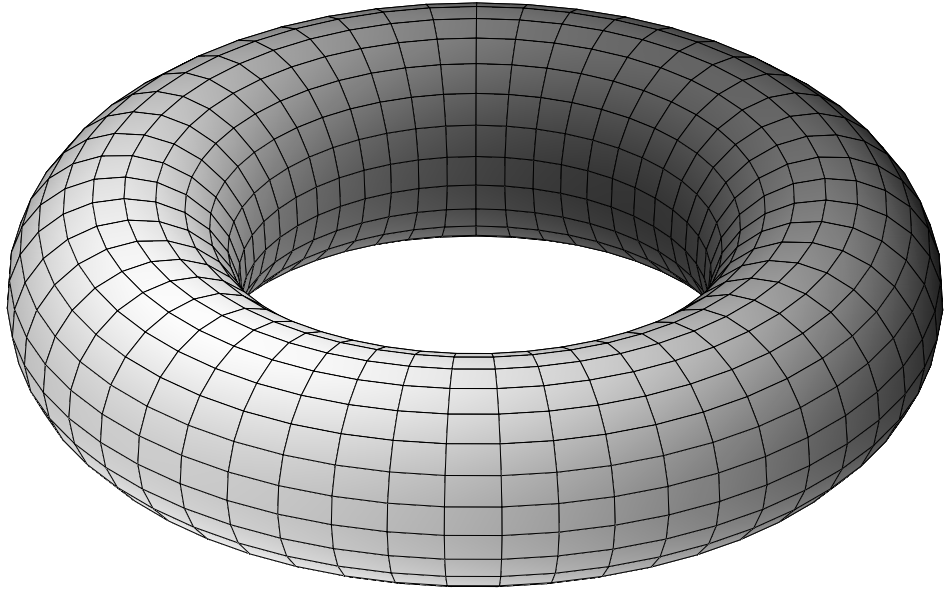

\paragraph{Regularization graph.}
We take the Cartesian product of two regularization graphs:
\begin{itemize}
    \item \emph{Week of year}.
    The regularization graph is a cycle graph with 52 nodes (one for each week of the year) 
    with edge weights $\gamma_{\mathrm{week}}$.
    \item \emph{Hour of day}.
    The regularization graph is a cycle graph with 24 nodes (one for each hour of the day) 
    with edge weights $\gamma_{\mathrm{hr}}$.
\end{itemize}
The Cartesian product of these two graphs is a torus,
illustrated in figure~\ref{fig:weather_reg_graph}.

\begin{figure}
  \centering
    \includegraphics[width=.9\textwidth]{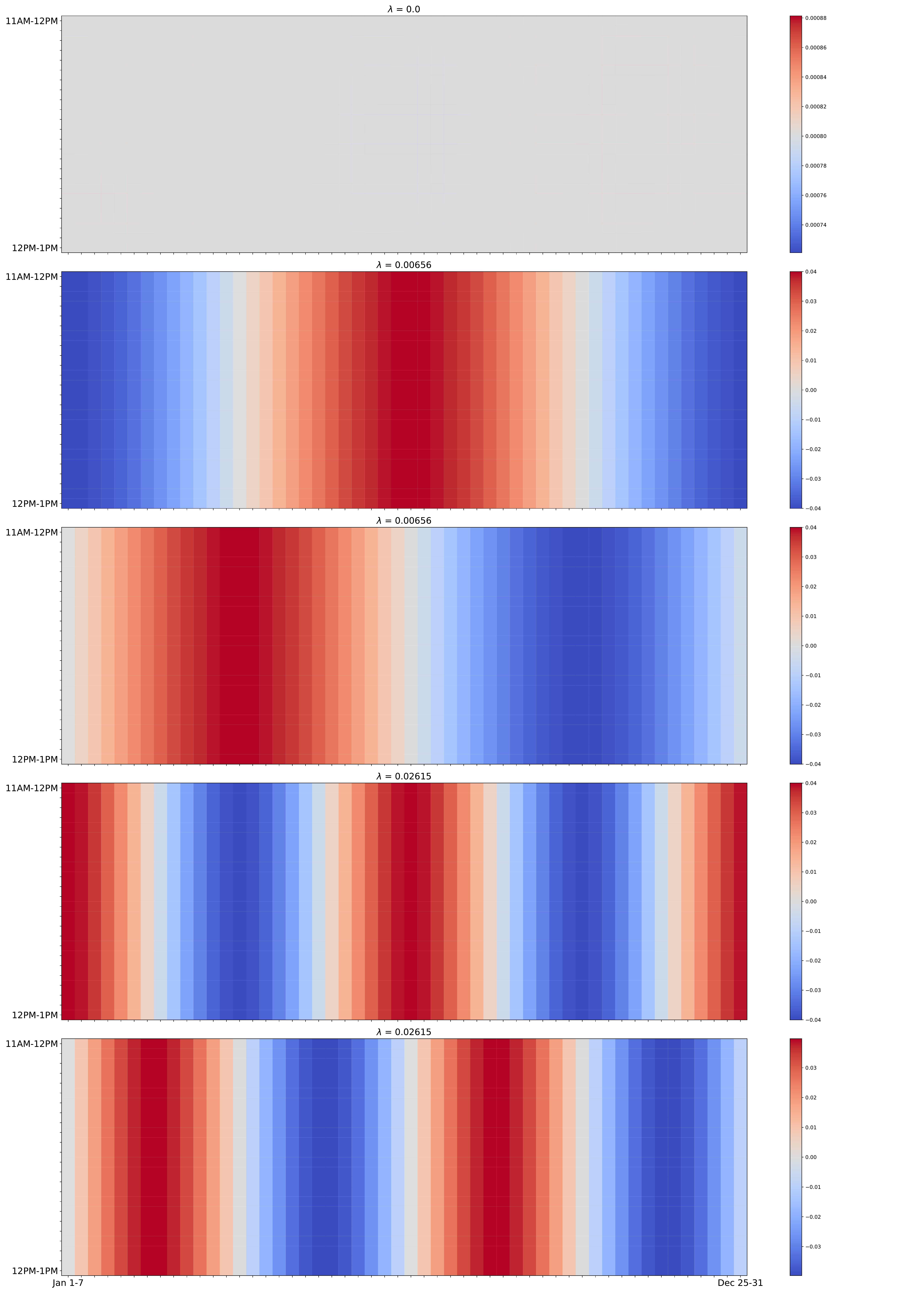}
      \caption{Heatmaps of the eigenvectors of the week/hour regularization 
      graph Laplacian corresponding to $\lambda_1, \ldots, \lambda_5$.}
  \label{fig:weather_eigen_heatmap1}
\end{figure}

\begin{figure}
  \centering
    \includegraphics[width=.9\textwidth]{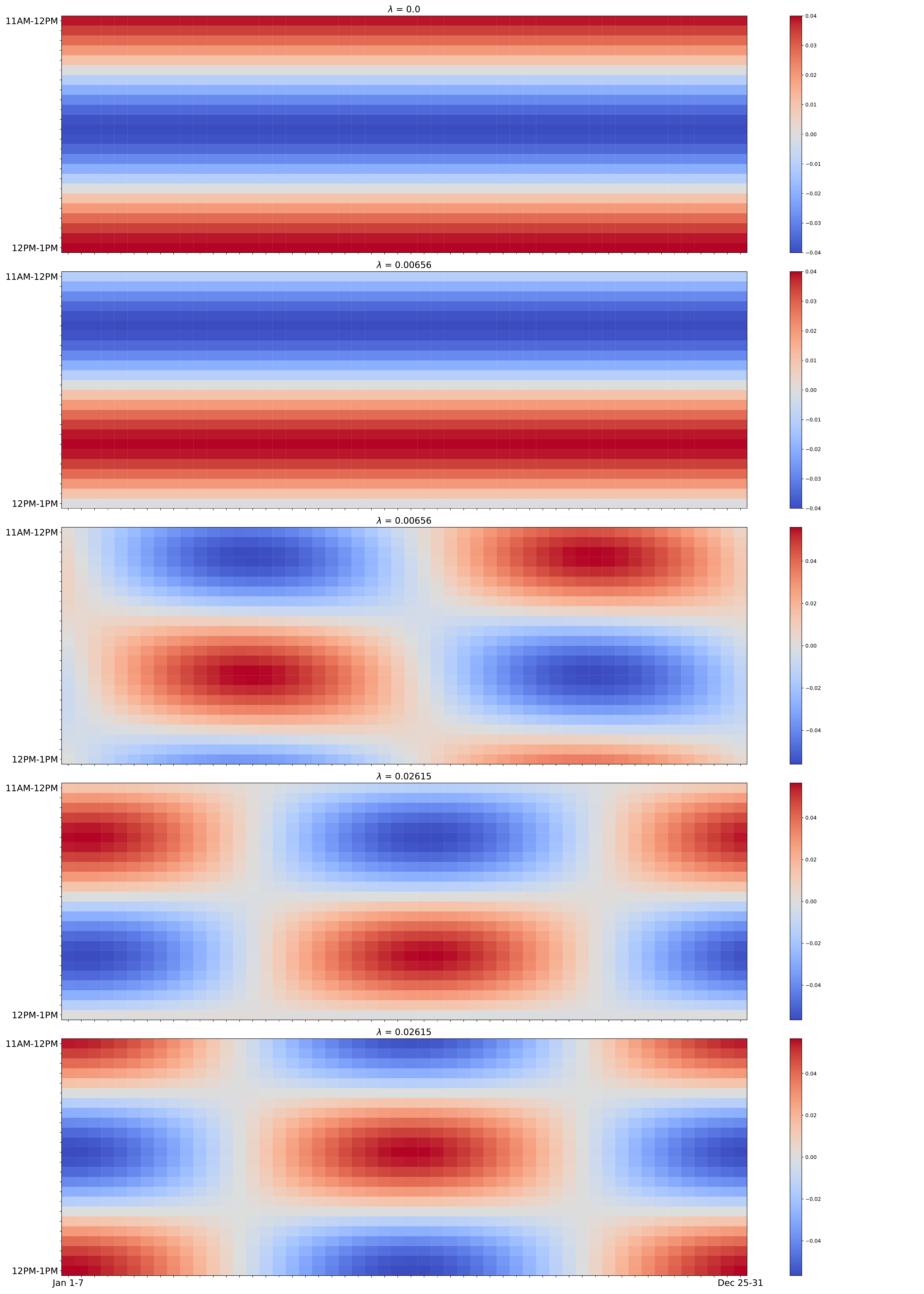}
      \caption{Heatmaps of the eigenvectors of the week/hour regularization 
      graph Laplacian corresponding to $\lambda_5, \ldots, \lambda_9$.}
  \label{fig:weather_eigen_heatmap2}
\end{figure}

This graph has $K=1248$ eigenvectors. 
The eigenvectors of this graph are given by
\BEAS
&& {\tilde s}_{i,j} / \|{\tilde s}_{i,j}\|_2\\
&& {\tilde u}_{i,j} / \|{\tilde u}_{i,j}\|_2\\
&& {\tilde v}_{i,j} / \|{\tilde v}_{i,j}\|_2\\
&& {\tilde w}_{i,j} / \|{\tilde w}_{i,j}\|_2,
\EEAS
where
\BEAS
&& s_{i,j} = \cos(\pi (i-1) u / 26) \otimes \sin(\pi (j-1) v / 12),\\ 
&& u_{i,j} = \cos(\pi (i-1) u / 26) \otimes \cos(\pi (j-1) v / 12),\\
&& v_{i,j} = \sin(\pi (i-1) u / 26) \otimes \sin(\pi (j-1) v / 12),\\
&& w_{i,j} = \sin(\pi (i-1) u / 26) \otimes \cos(\pi (j-1) v / 12),
\EEAS
for $i = 1, \ldots, 26$ and $j = 1, \ldots, 12$, $u = (0, \ldots, 51)$, $v = (0, \ldots, 23)$, 
and $\cos(\cdot)$ and $\sin(\cdot)$ are applied elementwise.
Figures \ref{fig:weather_eigen_heatmap1} and \ref{fig:weather_eigen_heatmap2} plots the bottom 10 
eigenvectors of the week/hour regularization graph Laplacian, with the particular 
week/hour edge weights $(\gamma_{\mathrm{week}},\gamma_{\mathrm{hr}}) = (.45,.55)$, 
sorted in increasing order corresponding to the bottom 10 eigenvalues of the Laplacian.

\paragraph{Results.}
For each of the fitting methods, we ran a crude hyper-parameter search over their hyper-parameters
and selected hyper-parameters that performed well over the validation set.
For the separate model, we used $\gamma_1=0.75$ and $\gamma_2=0.3$, and
for the common model, we used $\gamma_1=0.65$ and $\gamma_2=0.55$.
For the standard stratified model, we used $\gamma_1=0.05$, $\gamma_2=0.05$,
$\gamma_{\mathrm{week}} = 0.6$ and $\gamma_{\mathrm{hr}} = 0.5$.
For the eigen-stratified model, we used $\gamma_1=0.01$, $\gamma_2=0.001$,
$\gamma_{\mathrm{week}} = 0.45$ and $\gamma_{\mathrm{hr}} = 0.55$,
and $m=90$ (roughly 7\% of the $52\cdot24=1248$ eigenvectors).
We compare the ANLLs over the training, validation, and test datasets 
for the separate, common, standard stratified and eigen-stratified models
in table~\ref{tab:weather}.
\begin{table}
  \caption{Results for \S\ref{ss:weather}.}
    \vspace{.4em}
    \centering
    \begin{tabular}{llll}
      \toprule
      Model & Train ANLL & Validation ANLL & Test ANLL \\
      \midrule
      Separate & 0.186 & 0.447 & 0.448\\
      Common & 0.255 & 0.488 & 0.488\\
      Standard stratified & 0.172 & 0.393 & 0.394\\
      Eigen-stratified & 0.183 & 0.377 & 0.378\\
      \bottomrule
    \end{tabular}
    \label{tab:weather}
\end{table}
The validation and held-out test ANLLs of the eigen-stratified model were smaller 
than the respective ANLLs of every other model in table~\ref{tab:weather}, 
including those of the standard stratified model.

\begin{figure}
  \centering
    \includegraphics[width=\textwidth]{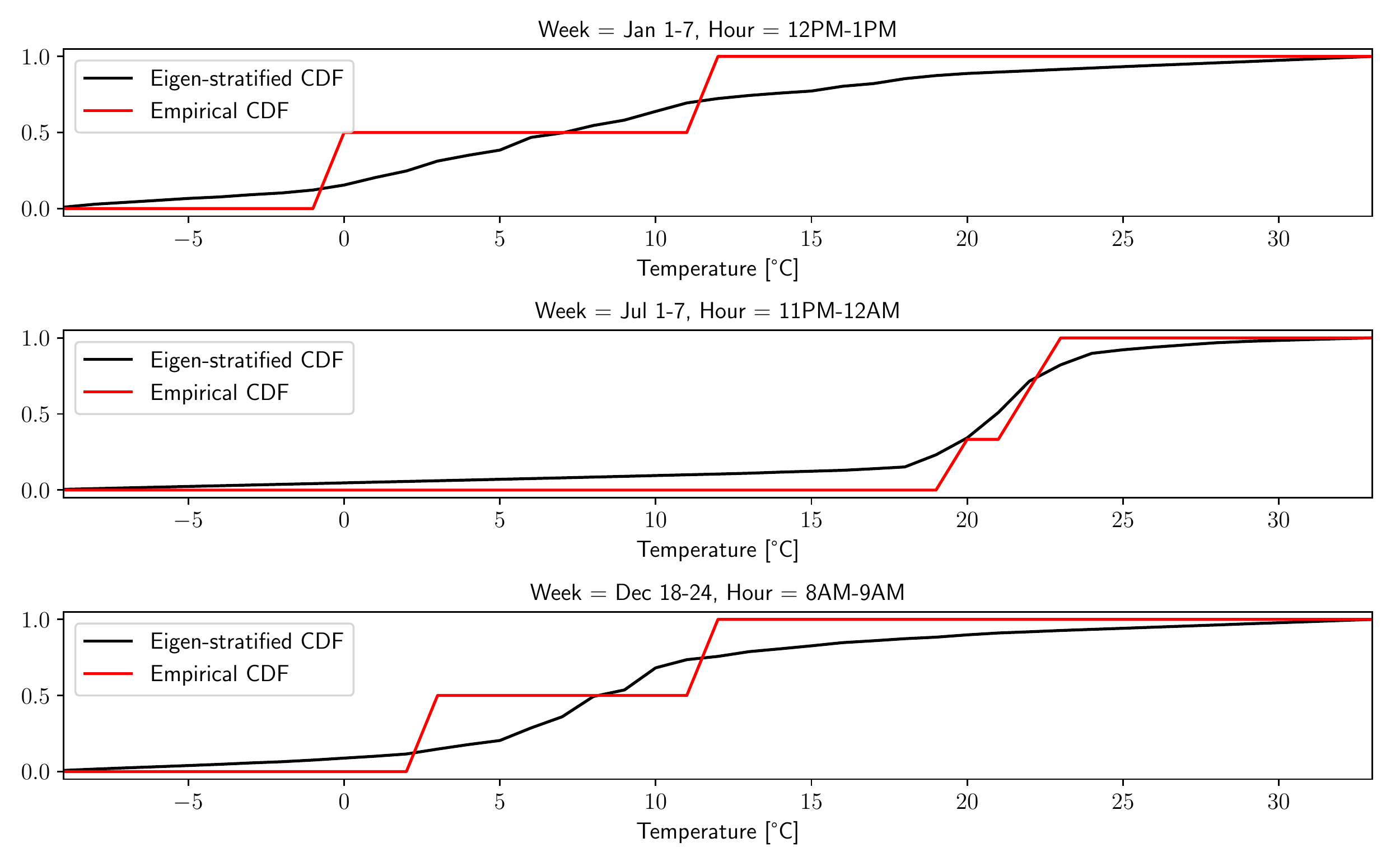}
      \caption{CDFs of various weeks of the year and hours of the day, 
      as given by the eigen-stratified model, along with their corresponding
      empirical CDFs.}
  \label{fig:weather_CDFs}
  \end{figure}

In figure~\ref{fig:weather_CDFs} we plot the cumulative distribution functions (CDFs) 
of temperature for week 1, hour 1; week 28, hour 12; and week 51, hour 21 
(which have 2, 3, and 2 empirical measurements in the training dataset, respectively),
for the eigen-stratified model, and for the test empirical data.

\begin{figure}
  \centering
    \includegraphics[width=\textwidth]{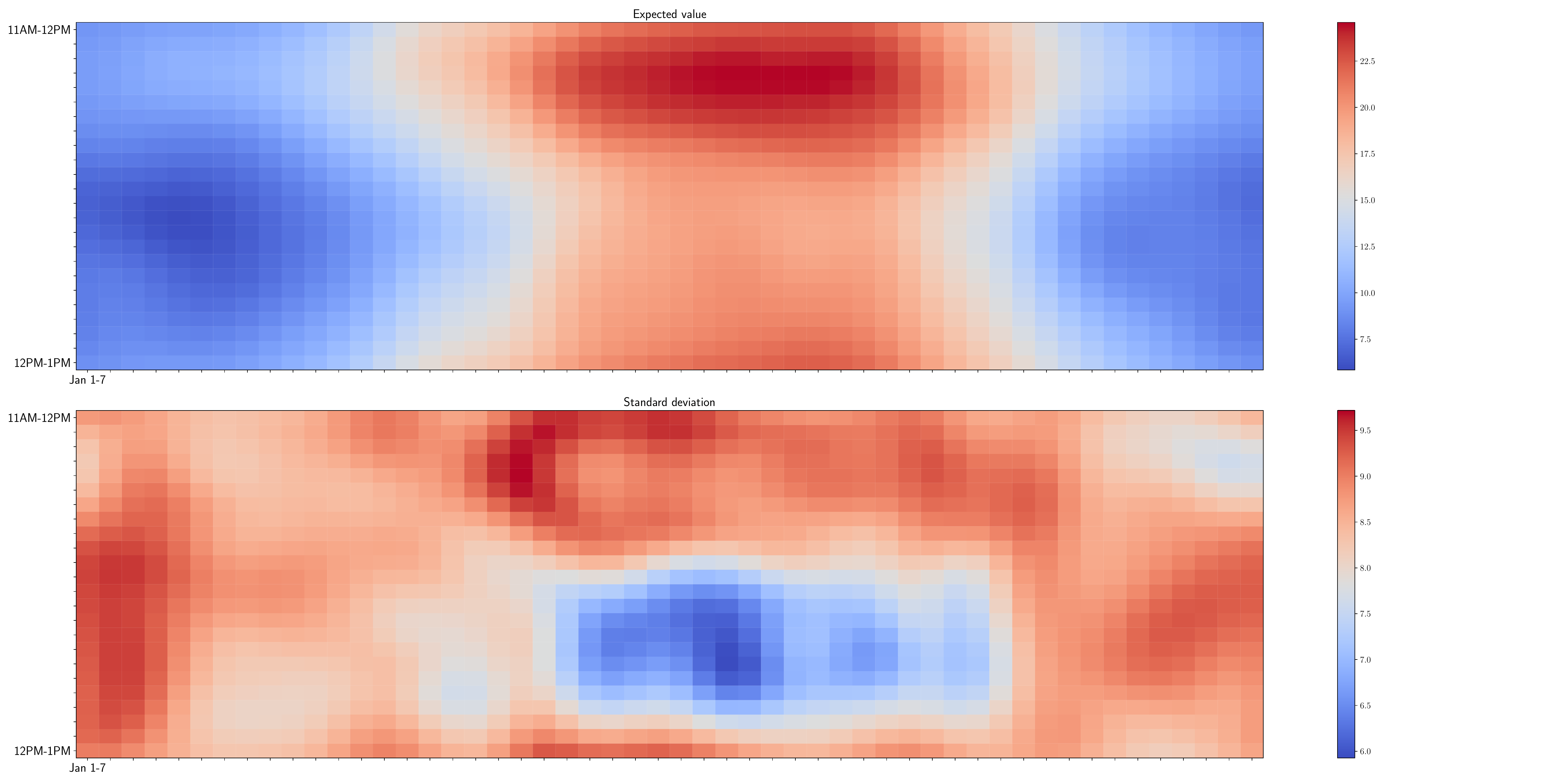}
      \caption{Heatmaps of the expected value (top) and standard deviation (bottom)
      of the distributions given by the eigen-stratified model.}
  \label{fig:weather_statistics}
  \end{figure}

In figure~\ref{fig:weather_statistics}, we plot heatmaps of the expected value
and standard deviation of the distributions given by the eigen-stratified model.
The statistics vary smoothly as hours of day and weeks of year vary.

\section*{Acknowledgements}
We gratefully acknowledge discussions with Shane Barratt and Peter Stoica, 
who provided us with useful suggestions that improved the paper.

\newpage
\bibliography{eigen_strat}

\newpage
\appendix

\section{Eigenvectors and eigenvalues of common graphs}\label{app:graphs}
The direct relation of a graph's structure to the eigenvalues and eigenvectors of its corresponding 
graph Laplacian is well-known~\cite{anderson1985eigenvalues}.
In some cases, mentioned below, we can find them analytically, especially when the graph 
has many symmetries.
The eigenvectors are given in normalized form (\ie, $\|q_k\|_2 = 1$.)
Outside of these common graphs, many other simple graphs can be analyzed 
analytically; see, \eg, \cite{brouwer2012graphs}.

\paragraph{A note on complex graphs.}
If a graph is complex, \ie, there is no analytical form for its graph Laplacian's 
eigenvalues and eigenvectors,
the bottom eigenvalues and eigenvectors of the Laplacian of a graph can be computed extremely 
efficiently by, \eg, the Lanczos algorithm or other more exotic methods.
We refer the reader to \cite{lanczos1950lanczos, ojalvo1970lanczos, paige1971eigvecs}
for these methods.

\begin{figure}
  \centering
  \begin{tikzpicture}
    \def \x {6}
    \def \minrad {2.5em}
        \node[shape=circle,draw=black, minimum size=\minrad] (1) at (-7,\x) {$1$};
        \node[shape=circle,draw=black, minimum size=\minrad] (2) at (-5,\x) {$2$};
        \node[shape=circle,draw=black, minimum size=\minrad] (3) at (-3,\x) {$3$};
        \node[shape=circle,draw=black, minimum size=\minrad] (4) at (-1,\x) {$4$};
        \node[shape=circle,draw=black, minimum size=\minrad] (5) at (1,\x) {$5$};
        \node[shape=circle,draw=black, minimum size=\minrad] (6) at (3,\x) {$6$}; 
        \node[shape=circle,draw=black, minimum size=\minrad] (7) at (5,\x) {$7$};
        \node[shape=circle,draw=black, minimum size=\minrad] (8) at (7,\x) {$8$};
        \draw[-] (1) -- (2);
        \draw[-] (2) -- (3);
        \draw[-] (3) -- (4);
        \draw[-] (4) -- (5);
        \draw[-] (5) -- (6);
        \draw[-] (6) -- (7);
        \draw[-] (7) -- (8);
    \end{tikzpicture}
    \caption{A path graph with 8 vertices and unit weights.}
    \label{fig:graph_path}
\end{figure}
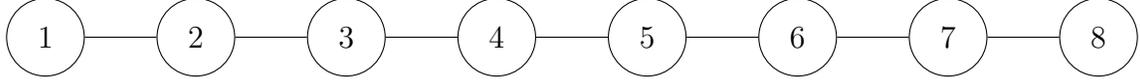

\paragraph{Path graph.}
A path or linear/chain graph is a graph whose vertices can be listed in order, 
with edges between adjacent vertices in that order. 
The first and last vertices only have one edge, whereas the other vertices have two edges.
Figure~\ref{fig:graph_path} shows a path graph with 8 vertices and unit weights.

Eigenvectors $q_1, \ldots, q_{K}$ of a path graph Laplacian with $K$ nodes and unit edge weights are given by
\BEAS
  q_k = \cos(\pi k v / K - \pi k / 2K)/\|\cos(\pi k v / K - \pi k / 2K)\|_2 \quad k = 0, \ldots, K-1,
\EEAS
where $v = (0, \ldots, K-1)$ and $\cos(\cdot)$ is applied elementwise.
The eigenvalues are $2-2\cos(\pi k/K), k = 0, \ldots, K-1$.

\begin{figure}
  \centering
  \begin{tikzpicture}

  \def \n {10}
  \def \radius {3cm}
  \def \margin {8} 
  
  \foreach \s in {1,...,\n}
  {
    \node[draw, circle] at ({360/\n * (\s - 1)}:\radius) {$\s$};
    \draw[-, >=latex] ({360/\n * (\s - 1)+\margin}:\radius) 
      arc ({360/\n * (\s - 1)+\margin}:{360/\n * (\s)-\margin}:\radius);
  }
  \end{tikzpicture}
  \caption{A cycle graph with 10 vertices and unit weights.}
  \label{fig:graph_cycle}
\end{figure}
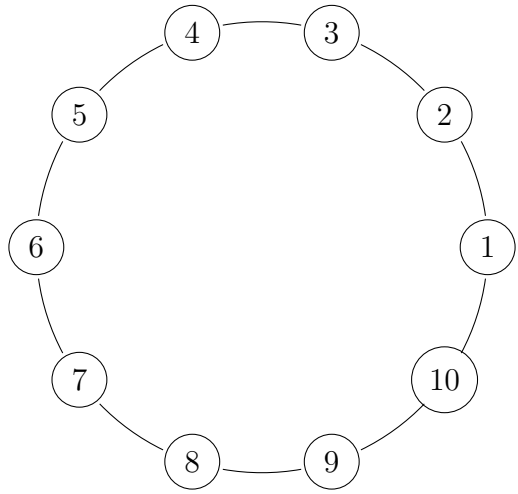

\paragraph{Cycle graph.}
A cycle graph or circular graph is a graph where the vertices are connected in a closed 
chain. 
Every node in a cycle graph has two edges.
Figure~\ref{fig:graph_cycle} shows a cycle graph with 10 vertices and unit weights.

Eigenvectors of a cycle graph Laplacian with $K$ nodes and unit weights are given by
\BEAS
  &\frac{1}{\sqrt{K}}\ones, k=0\\
  &\cos(2 \pi k v / K)/\|\cos(2 \pi k v / K)\|_2 
  \ \mathrm{and} \ 
  \sin(2 \pi k v / K)/\|\sin(2 \pi k v / K)\|_2,  \quad k = 1, \ldots, K/2,
\EEAS
where $v = (0, \ldots, K-1)$ and $\cos(\cdot)$ and $\sin(\cdot)$ are applied elementwise.
The eigenvalues are $2-2\cos(2\pi k/K), k = 0, \ldots, K-1$.

\begin{figure}
  \centering
  \newcommand{\stargraph}[2]{\begin{tikzpicture}
    \node[draw, circle, inner sep=0.25cm] at (360:0mm) (center) {0};
    \foreach \n in {1,...,#1}{
        \node[draw, circle, inner sep=0.25cm] at ({\n*360/#1}:#2cm) (n\n) {\n};
        \draw (center)--(n\n);
    }
\end{tikzpicture}}
\stargraph{9}{3}
\caption{A star graph with 10 vertices (9 outer vertices) and unit weights.}
\label{fig:graph_star}
\end{figure}
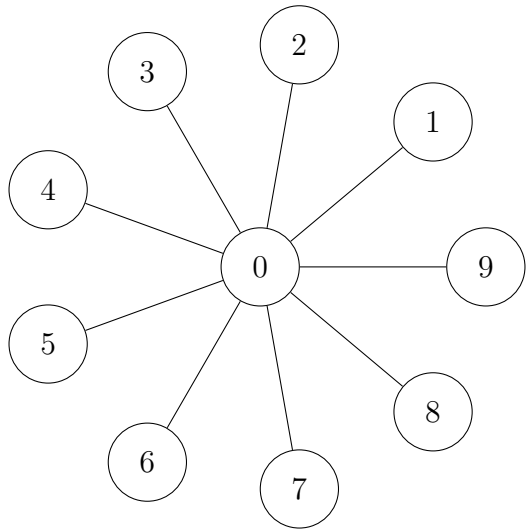

\paragraph{Star graph.}
A star graph is a graph where all of the vertices are only connected to one central
vertex.
Figure~\ref{fig:graph_star} shows an example of a star graph with 10 vertices (9 outer vertices) 
and unit weights.

Eigenvectors of a star graph with $K$ vertices (\ie, $K-1$ outer vertices) and unit
edge weights are given by
\BEAS
  & q_0 = \frac{1}{\sqrt{K}}\ones\\
  & q_k = \frac{1}{\sqrt{2}} (e_i - e_{i+1}), \quad 1 \leq i \leq K-2\\
  & q_{K-1} = \frac{1}{\sqrt{K(K-1)}}(K-1, -1, -1, \ldots, -1, -1),
\EEAS
where $e_i$ is the $i$th basis vector in $\reals^K$.
The smallest eigenvalue of this graph is zero, the largest eigenvalue is $K$,
and all other eigenvalues are 1.

\begin{figure}
  \centering
  \begin{tikzpicture}

  \def \n {10}
  \def \radius {3cm}
  \def \margin {8} 
  
  \node[draw, circle] (zero) at (0,0) {0};

  \foreach \s in {1,...,\n}
  {
    \node[draw, circle] (\s) at ({360/\n * (\s - 1)}:\radius) {$\s$};
    \draw[-, >=latex] ({360/\n * (\s - 1)+\margin}:\radius) 
      arc ({360/\n * (\s - 1)+\margin}:{360/\n * (\s)-\margin}:\radius);
    \draw[-] (\s) -- (zero);
  }
  \end{tikzpicture}
  \caption{A wheel graph with 11 vertices (10 peripheral vertices) and unit weights.}
  \label{fig:graph_wheel}
\end{figure}
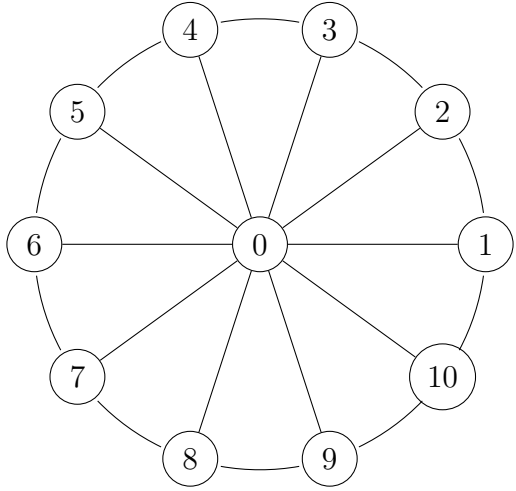

\paragraph{Wheel graph.}
A wheel graph with $K$ nodes consists of a center (\emph{hub}) vertex and a ring of $K-1$
peripheral vertices, each connected to the hub \cite{boyd2009fmmp}.
Figure~\ref{fig:graph_wheel} shows a wheel graph with 11 vertices (10 peripheral vertices) 
and unit weights.

Eigenvectors of a wheel graph with $K$ vertices (\ie, $K-1$ peripheral vertices) are given 
by~\cite{zhang2009wheelgraph}
\BEAS
  & q_0 = \frac{1}{\sqrt{K}}\ones\\
  & q_k = \sin(2 \pi k v / K) / \|\sin(2 \pi k v / K)\|_2, \quad 1 \leq i \leq K-2, i \text{ odd}\\
  & q_k = \cos(2 \pi k v / K) / \|\cos(2 \pi k v / K)\|_2, \quad 1 \leq i \leq K-2, i \text{ even}\\
  & q_{K-1} = \frac{1}{\sqrt{K(K-1)}}(K-1, -1, -1, \ldots, -1, -1),
\EEAS
where $v = (0, \ldots, K-1)$ and $\cos(\cdot)$ and $\sin(\cdot)$ are applied elementwise.
The smallest eigenvalue of the graph is zero, the largest eigenvalue is $K$, and
the middle eigenvalues are given by $3 - 2\cos(2 \pi i/(K-1)), i = 1, \ldots, (K-2)/2$,
with multiplicity 2 \cite{butler2008eigvals}.

\paragraph{Complete graph.}
A complete graph contains every possible edge; we assume here the edge weights are all one.
The first eigenvector of a complete graph Laplacian with $K$ nodes is $\frac{1}{\sqrt K} \ones$,
and the other $K-1$ eigenvectors are any orthonormal vectors that complete the basis.
The eigenvalues are 0 with multiplicity 1, and $K$ with multiplicity $K-1$.

\begin{figure}
  \centering
  \begin{tikzpicture}[transform shape,line width=0.2pt]
    \foreach \x in {1,...,8}{%
      \pgfmathparse{(\x-1)*45+floor(\x/9)*22.5}
      \node[draw,circle,inner sep=0.25cm] (N-\x) at (\pgfmathresult:4cm) [thick] {\x};
    } 
    \foreach \x [count=\xi from 1] in {2,...,8}{%
      \foreach \y in {\x,...,8}{%
          \path (N-\xi) edge[-] (N-\y);
    }
  }
  \end{tikzpicture}
  \caption{A complete graph with 8 vertices and unit weights.}
  \label{fig:graph_complete}
\end{figure}
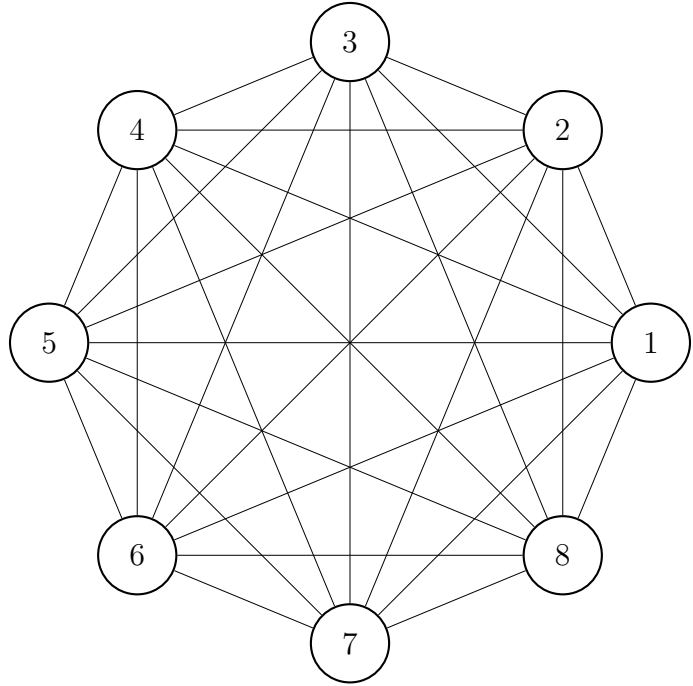

Figure~\ref{fig:graph_complete} shows an example of a complete graph with 8 vertices and unit weights.

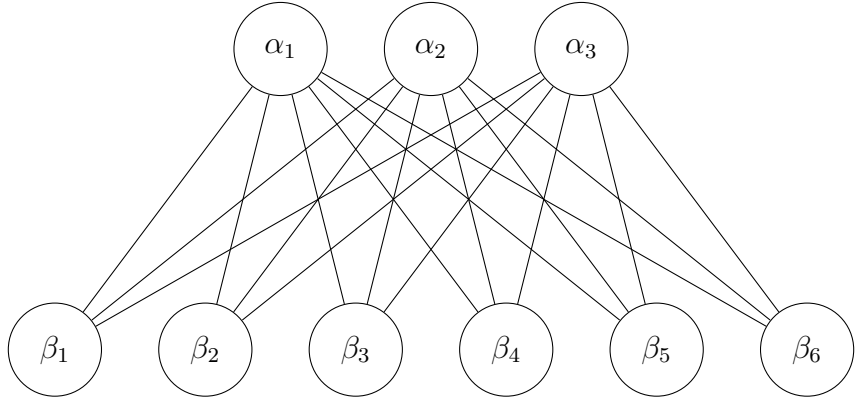
\begin{figure}
  \centering
  \begin{tikzpicture}
    \def \mh {2}
    \def \fh {6}
    \def \minrad {3em}
        \node[shape=circle,draw=black, minimum size=\minrad] (M1) at (0,\mh) {$\beta_1$};
        \node[shape=circle,draw=black, minimum size=\minrad] (M2) at (2,\mh) {$\beta_2$};
        \node[shape=circle,draw=black, minimum size=\minrad] (M3) at (4,\mh) {$\beta_3$};
        \node[shape=circle,draw=black, minimum size=\minrad] (M4) at (6,\mh) {$\beta_4$};
        \node[shape=circle,draw=black, minimum size=\minrad] (M5) at (8,\mh) {$\beta_5$};
        \node[shape=circle,draw=black, minimum size=\minrad] (M6) at (10,\mh) {$\beta_6$};

        \node[shape=circle,draw=black, minimum size=\minrad] (F1) at (3,\fh) {$\alpha_1$};
        \node[shape=circle,draw=black, minimum size=\minrad] (F2) at (5,\fh) {$\alpha_2$};
        \node[shape=circle,draw=black, minimum size=\minrad] (F3) at (7,\fh) {$\alpha_3$};

        \draw[-] (M1) -- (F1);
        \draw[-] (M1) -- (F2);
        \draw[-] (M1) -- (F3);
        \draw[-] (M2) -- (F1);
        \draw[-] (M2) -- (F2);
        \draw[-] (M2) -- (F3);
        \draw[-] (M3) -- (F1);
        \draw[-] (M3) -- (F2);
        \draw[-] (M3) -- (F3);  
        \draw[-] (M4) -- (F1);
        \draw[-] (M4) -- (F2);
        \draw[-] (M4) -- (F3);
        \draw[-] (M5) -- (F1);
        \draw[-] (M5) -- (F2);
        \draw[-] (M5) -- (F3);
        \draw[-] (M6) -- (F1);
        \draw[-] (M6) -- (F2);
        \draw[-] (M6) -- (F3);      
    \end{tikzpicture}
    \caption{A (3,6)-complete bipartite graph with unit weights.}
    \label{fig:graph_complete_bipartite}
\end{figure}

\paragraph{Complete bipartite graph.}
A bipartite graph is a graph whose vertices can be decomposed into two disjoint sets
such that no two vertices share an edge within a set.
A \emph{complete} bipartite graph is a bipartite graph such that every pair of vertices
in the two sets share an edge.
We denote a complete bipartite graph with $\alpha$ vertices on the first set and $\beta$
vertices on the second set as an $(\alpha, \beta)$-complete bipartite graph. 
We have that $\alpha+\beta=K$, and use the convention that $\alpha \leq \beta$.
Figure~\ref{fig:graph_complete_bipartite} illustrates an example of a complete bipartite graph with
$(\alpha, \beta)=(3,6)$ and unit weights.

Eigenvectors of an $(\alpha, \beta)$-complete bipartite graph with unit edge weights are 
given by~\cite{merris1998eigenvectors}:
\BEAS
  & q_0 = \frac{1}{\sqrt{K}}\ones\\
  & q_k = 
  \frac{1}{\sqrt{2}}(e_k - e_{k+1}), \quad 1 \leq k \leq \alpha-1\\
  & q_k = 
  \frac{1}{\sqrt{2}}(e_k - e_{k+1}), \quad \alpha \leq k \leq K-1\\
  & (q_{K-1})_i = \begin{cases}
    \frac{-\beta}{\sqrt{\alpha^2\beta + \beta^2\alpha}} & 1 \leq i \leq \alpha\\
    \frac{\alpha}{\sqrt{\alpha^2\beta + \beta^2\alpha}} & \alpha < i \leq K \\
  \end{cases}.
\EEAS
The eigenvalues are zero (multiplicity 1), 
$\alpha$ (multiplicity $\beta-1$), 
$\beta$ (multiplicity $\alpha-1$),
and $K=\alpha+\beta$ (multiplicity 1).

\paragraph{Scaling and products of graphs.}
We can find the eigenvectors and eigenvalues of the graph Laplacian of more 
complex graphs using some simple relationships.
First, if we scale the edge weights of a graph by $\alpha \geq 0$, the eigenvectors
remain the same, and the eigenvalues are scaled by $\alpha$.
Second, the eigenvectors of a Cartesian product of graph Laplacians are given by the Kronecker
products between the eigenvectors of each of the individual graph Laplacians; 
the eigenvalues consist of the sums of one eigenvalue from one graph and one
from the other.
This can be seen by noting that the Laplacian matrix of the Cartesian product of two graphs
with graph Laplacians $L_1 \in \reals^{P \times P}$ and $L_2 \in \reals^{Q \times Q}$ 
is given by
\[
  L = (L_1 \otimes I) + (I \otimes L_2),
\]
where $L$ is the Laplacian matrix of the Cartesian product of the two graphs.
With Cartesian products of graphs, we find
it convienent to index the eigenvalues and eigenvectors of the Laplacian 
by two indices, \ie, the eigenvalues may be denoted as $\lambda_{i,j}$ 
with corresponding eigenvector $q_{i,j}$ for $i = 0, \ldots, P-1$ and $j = 0, \ldots, Q-1$. 
(The eigenvalues will need to be sorted, as explained below.)

As an example, consider a graph which is the product of a chain graph with $P$ vertices, 
edge weights $\alpha^\text{ch}$ and eigenvalues $\lambda^{\text{ch}} \in \reals^P$; 
and a cycle graph with $Q$ vertices, edge weights $\alpha^\text{cy}$, and
eigenvalues $\lambda^{\text{ch}} \in \reals^P$.
The eigenvalues have the form 
\[
  \lambda^{\text{ch}}_i + \lambda^{\text{cy}}_j, 
  \quad i=0\ldots, P-1, \quad j=0, \ldots, Q-1,
\]
To find the $m$ smallest of these, we sort them.   
The order depends on the ratio of the edge weights,
$\alpha^\text{ch}/\alpha^\text{cy}$.

As a very specific example, take $P=4$ and $Q=5$, 
$\alpha^\text{ch} = 1$, and $\alpha^\text{cy} = 2$.
The eigenvalues of the chain and cycle graphs are
\[
  \lambda^{\text{ch}} = (0, 0.586, 2, 3.414),
  \quad
  \lambda^{\text{cy}} = (0, 2.764, 2.764, 7.236, 7.236).
\]
The bottom six eigenvalues of the Cartesian product of these two graphs 
are then
\[
0, 0.586, 2, 2.764, 2.764, 3.350.
\]

\end{document}